\pdfoutput=1
\documentclass[11pt]{article}

\usepackage[]{acl}

\usepackage{times}
\usepackage{latexsym}

\usepackage[T1]{fontenc}

\usepackage[utf8]{inputenc}

\usepackage{microtype}

\usepackage{inconsolata}

\usepackage{graphicx}

\usepackage[inline]{enumitem}

\usepackage{listings}
\usepackage{xcolor}
\usepackage{float} % For positioning the figure

\definecolor{keyword}{rgb}{0.2, 0.4, 0.6}
\definecolor{comment}{rgb}{0.0, 0.5, 0.0}
\definecolor{string}{rgb}{0.5, 0.0, 0.5}
\definecolor{function}{rgb}{0.0, 0.0, 1.0}
\definecolor{variable}{rgb}{0.0, 0.0, 0.0}
\definecolor{number}{rgb}{0.6, 0.3, 0.0}

\lstdefinestyle{mystyle}{
    commentstyle=\color{comment},          % Comment color
    keywordstyle=\color{keyword},          % Keyword color
    numberstyle=\tiny\color{gray},         % Line number color
    stringstyle=\color{keyword},           % String color
    identifierstyle=\color{variable},      % Variable color
    basicstyle=\ttfamily\footnotesize,     % Font style and size
    breakatwhitespace=false,               % Automatic breaks
    breaklines=true,                       % Automatic line breaking
    captionpos=b,                          % Caption position
    keepspaces=true,                       % Keeps spaces
    showspaces=false,                      % Do not show spaces
    showstringspaces=false,                % Do not show string spaces
    showtabs=false,                        % Do not show tabs
    tabsize=2,                             % Tab size
    frame=tb,                              % Single frame around code
    rulecolor=\color{black},               % Frame color
    morekeywords={self, def, class, return, from, import, as},  % Add keywords
    framesep=8pt,
}

\newcommand{\framework}{\textsc{TransformerRanker}}
\newcommand{\datasets}{\textsc{datasets}}
\newcommand{\transformers}{\textsc{transformers}}

\usepackage{adjustbox}
\usepackage{siunitx}[=v2]
\usepackage{booktabs}
\usepackage{tabularx}
\usepackage{amsmath}
\usepackage{makecell}

\usepackage{multirow}
\usepackage{dblfloatfix}

\renewcommand{\arraystretch}{1.25}

\title{\framework{}: A Tool for Efficiently Finding the Best-Suited Language Models for Downstream Classification Tasks}

\author{
  Lukas Garbas \footnotemark[1]
  \And
  Max Ploner \footnotemark[1]\footnotemark[2]\vspace*{2.5mm}\\
  \footnotemark[1] Humboldt-Universität zu Berlin\\
  \footnotemark[2] Science Of Intelligence \\
  \texttt{<firstname>.<lastname>@hu-berlin.de} \\ 
  \And
  Alan Akbik \footnotemark[1]\footnotemark[2]\\
}
  
\begin{document}
\maketitle
\begin{abstract}
Classification tasks in NLP are typically addressed by selecting a pre-trained language model (PLM) from a model hub, and fine-tuning it for the task at hand. However, given the very large number of PLMs that are currently available, a practical challenge is to determine which of them will perform best for a specific downstream task. 
With this paper, we introduce \framework{}, a lightweight library that efficiently ranks PLMs for classification tasks without the need for computationally costly fine-tuning. Our library implements current approaches for \textit{transferability estimation} (LogME, H-Score, kNN), in combination with layer aggregation options, which we empirically showed to yield state-of-the-art rankings of PLMs~\cite{garbas2024choose}.
We designed the interface to be lightweight and easy to use, allowing users to directly connect to the HuggingFace \transformers{} and \datasets{} libraries. Users need only select a downstream classification task and a list of PLMs to create a ranking of likely best-suited PLMs for their task. We make \framework{} available as a pip-installable open-source library.\footnote{\url{https://github.com/flairNLP/transformer-ranker}.
Released under the MIT License.} 
\end{abstract}

% Main content - Paper Sections
\section{Introduction}

There currently exists a multitude of pre-trained transformer language models (PLMs) that are
readily available \citep[e.g. through model hubs;][]{wolf-huggingface}. From a practical perspective, this raises the question of which PLM will perform best once fine-tuned for a specific downstream NLP task. However, 
since fine-tuning a PLM is both computationally costly and sensitive to hyperparameters (such as the learning rate used for fine-tuning), an exhaustive search of all models is infeasible. In practice, this restricts users to the exploration of only a small number of PLMs and may lead to the best-suited PLM for a particular task not being found. 

\begin{figure}
\vspace{9mm}
\centering
    \includegraphics[width=\linewidth]{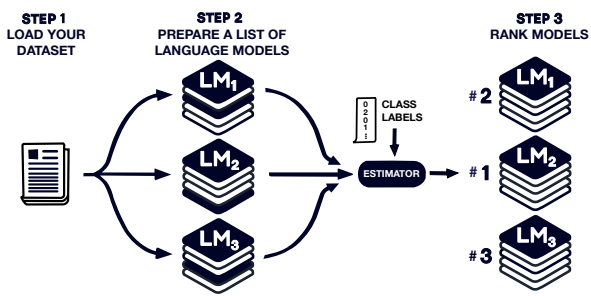}
    \vspace{-7mm}
    \caption{The three steps of \framework{}: (1)~The user selects a downstream classification task by selecting a dataset from HuggingFace \datasets{}. (2)~The user also selects a list of language models from HuggingFace \transformers{}. (3)~Using the selected estimator, the library returns a ranking of which PLMs are likely to perform best on the selected task.}
    \label{fig:process-illustration}
    \vspace{-4mm}
\end{figure}

To address this issue, prior work proposed methods for \textit{transferability estimation}. These methods avoid the high computational costs associated with fine-tuning a PLM by keeping the internal states frozen. Prominent examples of such methods include H-score \cite{baoInformationTheoreticApproachTransferability2019,hscore2023} and LogME \cite{logme2021}. In our prior work, we found that these methods can be improved by incorporating features from deeper layers in the estimation process~\cite{garbas2024choose}. We empirically showed that this yields better per-model estimates since, depending on the downstream task, different layers in the transformer model are best suited to provide features. Further, we showed that averaging across layers makes the selection process more robust against the different pre-training objectives used in each model, allowing a better comparison over a diverse set of PLMs.

\vspace{0.5mm}
\noindent 
\textbf{A library for transferability estimation.} With this paper, we present \framework{}, a Python library that enables users to leverage transferability estimation to identify the best-suited PLM for a downstream classification task. With \framework{}, we consolidate work in layer-wise analysis and transferability estimation methods into a single library and provide a three-step interface for ranking any transformer LMs available on the HuggingFace model hub (see Figure~\ref{fig:process-illustration}). Our goal is twofold: 

\begin{itemize}[left=0.82em]
    \setlength{\leftmargin}{0em}
    \item First, to give practitioners an easy-to-use method of using transferability estimation to select PLMs for their downstream tasks. To this end, we designed a simple interface that directly connects to the
HuggingFace \transformers{} and \datasets{} libraries. From these, users need only select a downstream classification task and a list of PLMs. Using our default settings, the library will output a ranking of likely best-suited PLMs for their task.

    \item Second, to assist researchers in the field of transferability estimation by having a single library that implements multiple state-of-the-art estimators and aggregation methods to combine and compare against.  
\end{itemize}

\framework{} supports NLP classification tasks of two main families: \textit{(1)} Text classification tasks such as question classification~\cite{voorhees2000overview}, sentiment analysis~\cite{socher2013recursive} or textual entailment~\cite{wang2018glue} in which a classification decision is made for an entire text (or text pair), \textit{(2)} sequence labeling tasks such as named entity recognition~\citep[NER;][]{sang-conll} and part-of-speech tagging~\cite{petrov2011universal} where classification decisions are made per-word.

Our library is built to rely solely on PyTorch~\cite{pytorch2019} and HuggingFace~\cite{wolf-huggingface} ecosystems and can be integrated as a fast model selection step in a larger NLP pipeline. We make \framework{} publicly available as a pip-installable open-source project. 
% ------------------------------------------------------------
% ---- TransformerRanker - Setup and three-step workflow -----
% ------------------------------------------------------------

\section{TransformerRanker}

We give an overview of \framework{} by first discussing the installation procedure (Section~\ref{sec:setup}) and the standard three-step workflow for ranking transformer models  (Section~\ref{sec:walkthrough}). We then discuss the implemented estimators  (Section~\ref{sec:estimators}) and layer aggregation methods (Section~\ref{sec:aggregation}).

\begin{lstlisting}[language=Python, caption={Full code example for finding the best-suited PLM for the \textsc{CoNLL-03} shared task on NER. We load the respective dataset and define a list of candidate PLMs. We run the ranker using default parameters to rank candidates according to their estimated suitability.}, label=code:full, float=t, belowskip=-0.6\baselineskip]
(*@\textbf{\# Step 1. Load Classification Dataset}@*)
dataset = load_dataset('conll2003')

(*@\textbf{\# Step 2. Create List of Candidate Models}@*)
language_models = [
    'bert-base-uncased',
    'xlm-roberta-large',
    'deberta-v3-base',
    ...
]

(*@\textbf{\# Step 3. Run Transformer Ranker}@*)
ranker = TransformerRanker(dataset)

result = ranker.run(
    language_models,
    batch_size=64
)

(*@\textbf{\# Inspect the result:}@*)
print(result)

(*@\textbf{\# Rank 1. 'deberta-v3-base'}@*)
(*@\textbf{\# Rank 2. 'xlm-roberta-large'}@*)
(*@\textbf{\# Rank 3. 'bert-base-uncased'}@*)
(*@\textbf{\# ...}@*)
\end{lstlisting}

\subsection{Setup}
\label{sec:setup}

\framework{} is implemented in Python and requires version 3.8 or higher. To install the library, run the following command: \texttt{pip install transformer-ranker}. This will download the latest version and all necessary dependencies, including \texttt{torch} and \texttt{transformers}. Alternatively, the library can be cloned directly from GitHub.

% ----------------------------------
% ---- 2.1 Three-step workflow -----
% ----------------------------------

\subsection{Three-Step Workflow}
\label{sec:walkthrough}

We illustrate the three-step workflow using a simple example in which we want to find the best-suited transformer model for the \textsc{CoNLL-03} shared task of English-language NER~\cite{sang-conll}. The full code needed to execute this search is shown in Listing~\ref{code:full}.

As the listing illustrates, our interface defines three main steps: First, we select the dataset that provides training and testing data for the task at hand, in this case \textsc{CoNLL-03}. Second, we create a list of language models to rank. Third, we initialize and run the ranker to obtain a ranking that indicates which models are best-suited for \textsc{CoNLL-03}.

\paragraph{Step 1: Load Your Dataset.}
To cover a large variety of existing NLP datasets, we provide support for those available in the HuggingFace \datasets{} library. The initial step involves loading an existing or custom dataset, by simply providing the corresponding dataset name (in this example \texttt{'conll03'}). 
As Listing~\ref{code:full} shows, no wrapper is introduced; the operations are performed directly on the structure provided by the \texttt{Dataset} and \texttt{DatasetDict} classes as returned by \datasets{}. 

\paragraph{Step 2: Create List of Candidate Models.}
The next step is to select the language models of interest. The list can include strings referring to transformers in the HuggingFace model hub or models already loaded using the \texttt{AutoModel} class from the \transformers{} library. In the example in Listing~\ref{code:full}, we simply define a set of string identifiers as our candidate models. 

Typically, users might chose candidate models by selecting the most popular (i.e., most downloaded models) on the HuggingFace model hub. We generally advise to search over models that were trained with different pre-training objectives and datasets in this list, as this can lead to significant performance differences. To help new users get started, we prepared two predefined lists of language models. The first contains smaller models (i.e. small and base models of popular PLMs), and the second contains large models. To load the list of smaller models, use the following code:

\begin{lstlisting}[language=Python, caption={}]
(*@\textbf{\# Step 2: Define model candidate list}@*)
language_models = popular_candidate_models('base')

\end{lstlisting}

Depending on the computational requirements of the project, we recommend new users to select either from smaller- or larger-size PLMs.

\vspace{1mm}

\paragraph{Step 3: Rank Models.} The final step is to initialize the \texttt{TransformerRanker} and run it. Initialization requires passing the dataset as a parameter. The dataset is preprocessed internally, retaining only the necessary columns (e.g., text and label). Next, the ranking process is executed by passing the language model list to the \texttt{run} method, together with optional hyperparameters (see Listing~\ref{code:full}). 

Two hyperparameters may be set to optimize the speed of the ranking approach: First, users must specify a \texttt{batch\_size} which indicates how many data points are embedded with a single forward pass. Depending on the available GPU memory, this parameter can be increased for quicker processing. Second, users may provide an optional parameter \texttt{dataset\_downsample} which downsamples the data to a percentage of its original size. This may yield significant speedups in model ranking, as we experimentally show in Section~\ref{sec:runtime}. We provide an example of how to downsample the dataset to 20\% of its size in Listing~\ref{code:alt}. 

Additionally, two hyperparameters may be set to configure the transferability estimation approach. By default, H-score is used as the estimator and \textit{layer mean} as the layer aggregation method, a combination we found to yield the best rankings~\cite{garbas2024choose}. We thus advise most users to use these default settings. However, it is also possible to exchange estimators and aggregation methods with other options discussed in Sections~\ref{sec:estimators} and~\ref{sec:aggregation},  using the \texttt{estimator} and \texttt{layer\_aggregation} parameters. An example of instead using \textit{kNN} with \textit{best layer} is illustrated in Listing~\ref{code:alt}. 

\begin{lstlisting}[language=Python, caption={Alternative configuration of Step 3 from Listing~\ref{code:full}: The dataset is downsampled to 20\% of its original size by passing \texttt{dataset\_downsample=0.2}. An alternative \texttt{estimator} and \texttt{layer\_aggregation} method are used for ranking.}, label=code:alt, float=t, belowskip=-1.5\baselineskip]
(*@\textbf{\# Initialize the ranker a downsampled dataset}@*)
ranker = TransformerRanker(
    dataset,
    dataset_downsample=0.2
)

(*@\textbf{\# Set estimator and layer aggregator and run}@*)
result = ranker.run(
    language_models,
    estimator='knn',
    layer_aggregation='bestlayer',
    batch_size=64
)

\end{lstlisting}

\paragraph{Inspecting the Result.} The ranking is compiled into a \texttt{Results} object that may be printed to inspect the ranking of language models and their transferability scores. An example ranking produced for 20 language models on the \textsc{CoNLL-03} shared task is shown in Figure~\ref{fig:output}.

The model with the best estimate will be at the top of the displayed list. Practitioners can use this information to select the most promising PLMs and proceed to fine-tune them using another framework such as \textsc{Flair} \cite{akbik2019flair} or \textsc{Transformers} \cite{wolf-etal-2020-transformers}. 

\begin{figure*}[thp] % the figure provides the caption
  \vspace{-4mm}
\centering          % which should be centered
\begin{tabular}{c}  % the tabular makes the listing as small as possible and centers it
\midrule
\begin{lstlisting}[language=Python, label={gtt_c_ausgabe}, frame=none]
Rank 1.   'google/electra-large-discriminator':       2.6960
Rank 2.   'microsoft/deberta-v3-base':                2.6257
Rank 3.   'bert-large-uncased':                       2.6165
Rank 4.   'xlm-roberta-large':                        2.6120
Rank 5.   'microsoft/mpnet-base':                     2.5957
Rank 6.   'distilbert-base-uncased':                  2.5742
Rank 7.   'cardiffnlp/twitter-roberta-base':          2.5723
Rank 8.   'microsoft/mdeberta-v3-base':               2.5594
Rank 9.   'google/electra-base-discriminator':        2.5586
Rank 10.  'roberta-base':                             2.5542
Rank 11.  'typeform/distilroberta-base-v2':           2.5445
Rank 12.  'bert-base-uncased':                        2.5413
Rank 13.  'sentence-transformers/all-mpnet-base-v2':  2.5244
Rank 14.  'xlm-roberta-base':                         2.4445
Rank 15.  'sentence-transformers/all-MiniLM-L12-v2':  2.1480
Rank 16.  'google/electra-small-discriminator':       1.9119
Rank 17.  'SpanBERT/spanbert-base-cased':             1.6703
Rank 18.  'allenai/scibert_scivocab_cased':           1.5767
Rank 19.  'dmis-lab/biobert-base-cased-v1.2':         1.5231
Rank 20.  'emilyalsentzer/Bio_ClinicalBERT':          1.3159
\end{lstlisting}
\end{tabular}
\caption{A ranking of 20 language models produced by \framework{} for the \textsc{CoNLL-03} shared task data. The output is ordered by rank, with the estimated best-suited model at the top of the list. For each model, the H-score is printed in the third column. Using these results, a user may exclude the lower-ranked models to only focus on the top-ranked models for further exploration.}
  \vspace{-4mm}
\label{fig:output}
\end{figure*}

% -------------------------
% ---- 2.2 Estimators -----
% -------------------------

\subsection{Estimators}
\label{sec:estimators}

Each PLM is scored by an estimator to assess its suitability for a classification task. We extract hidden states and pool them into word- or sentence-level embeddings as described in Appendix~\ref{appendix:embedder}. These embeddings are kept on GPU by default, or moved to CPU if specified. Once stored, they are scored using one of the available estimators:

\paragraph{Nearest Neighbour} Embedding suitability can be evaluated through a nearest neighbor perspective. We adapt the kNN algorithm~\cite{cover1967nearest} as an estimator, applying it to the entire dataset. A pairwise distance matrix is calculated between all data points, with diagonal values excluded to avoid self-distances during the top-k search. To manage memory, we handle distance calculation and top-k search in batches, eliminating the need to store a large distance matrix. In our implementation, the
computation is done using PyTorch, with batched distance calculations parallelized on a GPU. The kNN estimator has one hyperparameter $k$.

\paragraph{LogME} \citet{logme2021} assumes a linear relationship between embeddings and labels and proposes an algorithm to estimate the Bayesian evidence. 
Instead of training a linear layer to find the best set of weights, which can be prone to overfitting, LogME computes a closed-form solution for the model evidence by marginalizing over the weights.
Through evidence maximization, two key parameters, $\alpha$ and $\beta$, are introduced, controlling the regularization strength and noise.
The two parameters are updated using fixed-point iteration to maximize the marginal likelihood. 
LogME is computed by performing Singular Value Decomposition (SVD) on the feature matrix, followed by iterative maximization of $\alpha$ and $\beta$ for each class. We implement this estimator in PyTorch for GPU speedup during the SVD and matrix multiplications required in the fixed-point iteration.

\paragraph{H-Score} \cite{baoInformationTheoreticApproachTransferability2019} assess embedding suitability by evaluating how well they distinguish between classes based on their variability. The intuition is that well-suited representations should exhibit low within-class variation and high between-class differences, which is captured by using class means in the covariance calculation. By relying solely on covariance matrices, H-Score measures class separability without the need for training or iterative methods. \citet{hscore2023} addressed practical issues with covariance estimation in high-dimensional settings by proposing a shrinkage-based version that regularizes the covariance matrix and uses pseudo-inversion for improved stability. We use PyTorch and GPU to accelerate matrix operations, particularly for covariance computations and pseudo-inversion.

\begin{table*}[t!]
  \vspace{-2mm}
\centering
\setlength{\tabcolsep}{10pt}
\begin{tabularx}{\textwidth}{Xccc}  % Adjust the width as needed
\toprule
Ranking Method & \makecell{Sentence-level} & \makecell{Word-level} & Average \\
\hline
Linear Probing\textsubscript{ layer mean} & 0.89 \textsubscript{ $\tau$ = 0.76} & 0.78 \textsubscript{ $\tau$ = 0.66} & 0.84 \textsubscript{ $\tau$ = 0.71}\\ 
\hline
kNN\textsubscript{ best layer} & 0.68 \textsubscript{ $\tau$ = 0.63} & 0.70 \textsubscript{ $\tau$ = 0.55} & 0.68 \textsubscript{ $\tau$ = 0.55}\\ 
H-score\textsubscript{ layer mean} & \textbf{0.91} \textsubscript{ $\tau$ = \textbf{0.84}} & \textbf{0.85} \textsubscript{ $\tau$ = 0.64} & \textbf{0.88} \textsubscript{ $\tau$ = 0.74}\\ 
LogME\textsubscript{ layer mean} & 0.90 \textsubscript{ $\tau$ = 0.80} & 0.82 \textsubscript{ $\tau$ = \textbf{0.70}} & 0.86 \textsubscript{ $\tau$ = \textbf{0.75}}\\ 
\bottomrule
\end{tabularx}
\caption{Pearson’s $\rho$ correlation and weighted Kendall’s $\tau$ between the estimated scores and fine-tuning scores on three sentence and three per word classification tasks. The results are summarized from \citet{garbas2024choose} to illustrate the expected accuracy.
}
\label{citation-guide}
  \vspace{-4mm}
\end{table*}

% ------------------------------
% ---- 2.3 Layer Selection -----
% ------------------------------

\subsection{Layer Aggregation Strategies}
\label{sec:aggregation}

Transferability estimation methods can be improved by including hidden states from deeper layers in the estimation process. The library supports three options: the \textit{last layer}, which uses the hidden states of the final layer of the language model, a common practice when extracting features from a pre-trained model; the \textit{layer mean}, which averages the hidden states of all layers into a single representation; and the \textit{best layer}, which scores all layers of the language model separately and selects the one with the highest transferability score. Intuitively, by setting aggregation to \textit{best layer}, models are ranked by the scores of their task-specific layers.

The first two methods require a single estimation for a model, while \textit{best layer} requires an estimation for each layer. Identifying the layer with the best estimate is useful not only in model ranking but also for detecting the most suited layer within a model. Task-specialty of layers within a single model was studied by \citet{xie2022hiddenstatevariabilitypretrained} where a similar metric using the within- and between class variability was used. It was shown that the proposed metric correlates well to training a linear layer and correctly identifies well-suited layers for classification tasks.
\vspace{-1mm}
\section{Experiments}
\vspace{-2mm}

In this section, we present the estimation quality of different ranking approaches and evaluate how downsampling the data affects runtime and estimation quality. We further conduct experiments to measure the runtime of our GPU-accelerated implementations. Finally, we present an example
practical application of \framework{}.

% -----------------------------------------------------
% ---- 3.1 Accuracy of Transferability Estimation -----
% -----------------------------------------------------

\subsection{Accuracy of Transferability Estimation}

Table~\ref{citation-guide} summarizes the findings of~\citet{garbas2024choose}, where the rankings of various estimators and layer aggregation methods were compared against a gold ranking of models obtained by full fine-tuning and hyperparameter selection. To compare rankings, we used Pearson’s $\rho$ and weighted Kendall’s $\tau$ correlations. We present averaged scores over 6 tasks from two task families: word- and sentence-level classification. 

H-score outperformed LogME in Pearson's $\rho$ with scores of 0.88 and 0.86, respectively. In terms of weighted Kendall’s $\tau$, H-score and LogME showed similar performance, with average scores of 0.75 and 0.74, where $\tau$ for LogME was higher on word-level tasks. For both metrics, the highest scores were obtained using the \textit{layer mean} aggregation method. Consistent with their original works, both H-score and LogME outperform linear probing and kNN approaches.

Accordingly, we set H-score with \textit{layer mean} as the default estimator in \framework{}. Note, however, that other estimators may benefit from different layer aggregation strategies. For example, kNN achieves its best performance with the \textit{best layer} strategy (see Table~\ref{citation-guide}). For this reason, we give users the option of selecting both the estimator and the layer aggregator. 

% ---------------------------------------------
% ---- 3.2 Effect of Dataset Downsampling -----
% ---------------------------------------------

\subsection{Effect of Dataset Downsampling}
\label{sec:downsampling}

\begin{figure}[t]
    \vspace{1mm}
    \includegraphics[width=\linewidth]{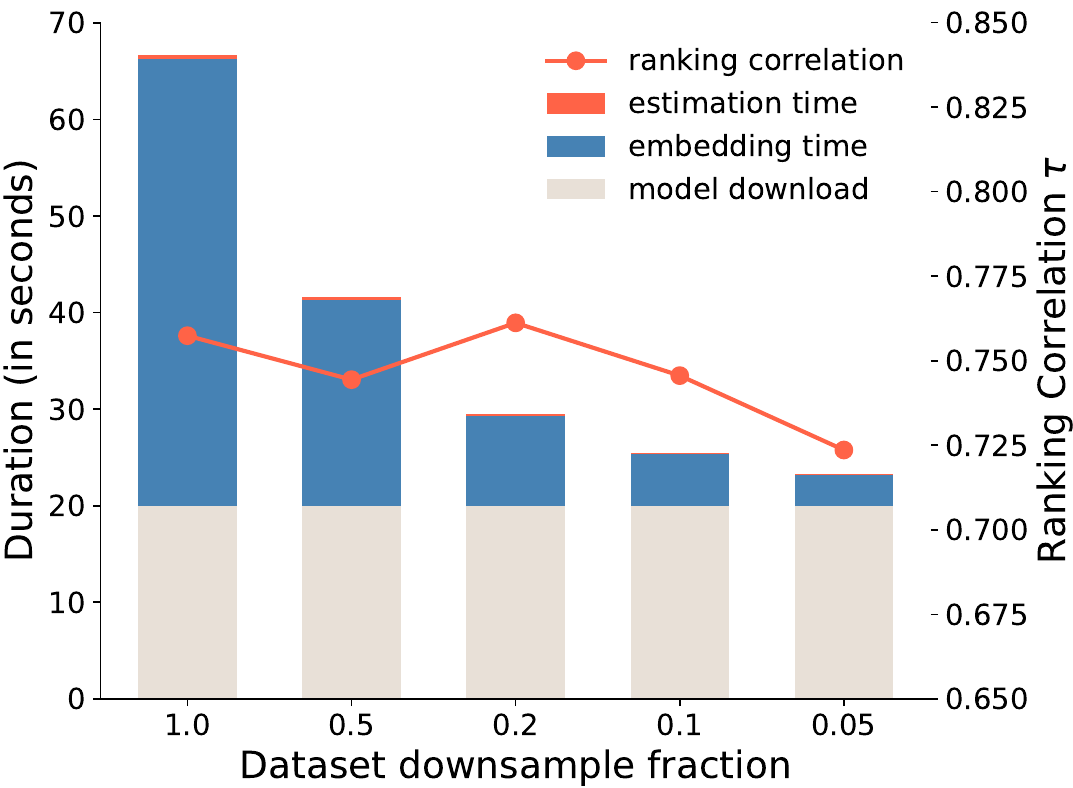}
  \vspace{-3mm}
    \caption{Time taken to estimate a single model. We include the download time (20 seconds), and report runtimes for different downsample splits of the \textsc{CoNLL-03} dataset. The plot also shows how the ranking correlation changes with different dataset splits. We used the default parameters of \textit{layer mean} with the h-score estimator. Estimation was done using a batch size of 64 on a single Nvidia A100 (80GB) GPU.}
    \label{fig:runtime-bars}
    \vspace{-4mm}
\end{figure}

Figure~\ref{fig:runtime-bars} shows the time required to estimate the transferability of a large PLM (DeBERTa-large~\cite{deberta2020}) using a dataset with 22K sentences. Estimating using the full dataset takes about one minute. We break the runtime into three parts: model download, dataset embedding, and estimating embedding suitability. The model download only occurs on the first run if it is not already stored locally. The most time is spent on embedding the complete dataset. 

Since downsampling can substantially speed up the embedding, we conduct an experiment in which we downsample the CoNLL-03 dataset to 50\%, 20\%, 10\%, and 5\% respectively.
As Figure~\ref{fig:runtime-bars} illustrates, the dataset can be sampled down substantially without a reduction in ranking correlation.

% -----------------------
% ---- 3.2 Runtimes -----
% -----------------------

\subsection{Runtime of Estimators}
\label{sec:runtime}

We evaluate our implementations of the kNN, H-score, and LogME estimators through runtime experiments in which we use 100,000 embeddings with a hidden size of 1024. Unlike the original implementations, the estimators in \framework{} can employ GPU acceleration. We thus measure the runtime on both devices.

Table \ref{table:runtime-devices} shows that H-score is faster to compute than LogME, which is consistent with the findings in the original work~\cite{hscore2023}. Further, we find that our implementation of kNN is by far the slowest of the three estimators. We also note significant speed-ups of GPU acceleration, amounting to about an order of magnitude faster computation on GPU. 

\begin{table}[!t]
 % \vspace{-2mm}
\centering
\small
\setlength{\tabcolsep}{4pt} % Adjust column separation
\renewcommand{\arraystretch}{1.2} % Adjust row separation
\begin{tabularx}{\linewidth}{X S[table-format=2.2] S[table-format=1.2] S[table-format=1.2]}
\toprule
 & \multicolumn{3}{l}{\textbf{Estimator}} \\
\textbf{Device} &      {kNN} &  {H-score} &  {LogME} \\
\midrule
CPU  & 17.76 s & 1.44 s & 3.82 s \\
GPU  &  3.43 s & 0.20 s & 0.77 s \\
\bottomrule
\end{tabularx}
\caption{Runtime comparison for estimators on CPU and GPU. Estimation runtimes are reported for 100,000 embeddings, excluding the time for embedding retrieval (i.e., the forward pass through a transformer). The CPU used was an Intel Xeon Gold 6134 @ 3.20GHz, and the GPU was an NVIDIA A100.}
\label{table:runtime-devices}
  \vspace{-5mm}
\end{table}

% -------------------
% ---- 3.2 Demo -----
% -------------------

\subsection{Demonstration}\label{sec:practical-example}

To provide an example of how \framework{} may be used practically, we choose the \textsc{GermEval18} task of fine-grained classification of offensive language in German \citep{wiegand2018overview}. Our baseline is the default approach a practitioner might take: Simply using the most popular PLMs for German on HuggingFace. In this case, our baselines are \textit{(1)} \texttt{bert-base-german-cased}, the most frequently downloaded German model, and \textit{(2)} \texttt{xlm-roberta-base}, the most frequently downloaded multilingual model. 

We use \framework{} to find one additional promising model from our predefined list of base models\footnote{We only select the highest-ranking model though in practice one might consider the top 3 highest-ranking models.}. We downsample \textsc{GermEval18} to 50\% for model selection.
The produced ranking of all models is presented in Figure~\ref{fig:germeval18-output}: Here, \framework{} estimates \texttt{mdeberta-base} to be the most suited model for this task.

\begin{figure}[t] 
\vspace{-0.5mm}
\centering      
\begin{tabular}{c}  
\midrule
\begin{lstlisting}[language=Python, label={gtt_c_ausgabe}, frame=none, columns=fixed]
Rank 1.  'mdeberta-v3-base':   0.9740
Rank 2.  'twhin-bert-base':    0.9426
Rank 3.  'deberta-v3-base':    0.9112
Rank 4.  'xlm-roberta-base':   0.8884
Rank 5.  'bert-german-cased':  0.8633
Rank 6.  'electra-base':       0.8064
Rank 7.  'roberta-base':       0.7970
Rank 8.  'bert-base-cased':    0.7943
Rank 9.  'PharmBERT-cased':    0.7679
Rank 10. 'biobert-base-cased': 0.7657
Rank 11. 'distilbert-base':    0.7504
Rank 12. 'distilroberta-base': 0.7482
Rank 13. 'all-mpnet-base-v2':  0.7427
Rank 14. 'spanbert-base':      0.7348
Rank 15. 'scideberta':         0.7173
Rank 16. 'all-MiniLM-L12-v2':  0.5025
Rank 17. 'electra-small':      0.3330
\end{lstlisting}
\end{tabular}
\vspace{-1mm}
\caption{\textsc{GermEval18} ranking result with H-scores for models from the predefined list of smaller PLMs.}
\label{fig:germeval18-output}
\end{figure}

We then fine-tune the identified model (\texttt{mdeberta-base}) and compare it to the two baseline models (\texttt{bert-base-german-cased}, \texttt{xlm-roberta-base}) w.r.t.~the accuracy of respective fine-tuned models. As Table~\ref{tab:demonstration-experiment} shows, we find that \texttt{mdeberta-v3-base} outperforms the two baseline models significantly. This example illustrates the potential value of using automated transferability estimation to identify well-suited PLMs for downstream tasks. 

\begin{table}[t]
    \centering
    %\small % Reduce the font size
    \setlength{\tabcolsep}{4pt} % Adjust column separation
    \begin{tabular}{lcc}
        \toprule
        \textbf{Model} & \textbf{Dev} & \textbf{Test} \\
        \midrule
        \texttt{mdeberta-base} & \textbf{78.41} \small{$\pm$ 0.97} & \textbf{75.86} \small{$\pm$ 0.54} \\ 
        \texttt{xlm-roberta-base} & 77.56 \small{$\pm$ 1.20} & 74.34 \small{$\pm$ 0.13} \\
        \texttt{bert-base-german} & 77.31 \small{$\pm$ 0.43} & 73.55 \small{$\pm$ 0.19} \\
        \bottomrule
    \end{tabular}
    \caption{Fine-tuned scores on the GermEval18 dataset for fine-grained classification of offensive language in German. We compare the two most popular models for text classification in German to models that were ranked highest by \framework{}.}
    \label{tab:demonstration-experiment}
    \vspace{-2mm}
\end{table}

\section{Conclusion and Outlook}

In this paper, we introduced \framework{}, a library that aids researchers and practitioners in systematically selecting well-suited PLMs for their downstream NLP tasks. In a fraction of the time, the underlying approach yields a ranking of candidate models that strongly correlates with models ranked by their true downstream performance. 

Future work will extend the scope of the addressed downstream tasks and incorporate new methods for transferability estimation. For instance, we are currently adding experimental support for regression tasks (limited to LogME and kNN estimators). Further, we invite the community to use \framework{} in their research and add other estimators such as Gaussian Bhattacharyya Coefficient \cite{pandy-etal-2021-bhattacharyya} and new layer aggregation methods.
% ----------------------------------------------------------
% ---- Limitations, Ethics Statement, Acknowledgements -----
% ----------------------------------------------------------

\section*{Limitations}

\framework{} requires users to provide a list of candidate models, which it automatically ranks with regards to their suitability on a downstream task. However, since this candidate list must be provided by the user, this means that model exploration may be limited to models already known to the user. One potential solution for future work may be to extend the framework with a component that generates candidates based on the models' metadata.

Another limitation is that, in order to calculate the embeddings of task samples, each candidate model must be downloaded or already present in the cache. Loading a large number of models can create significant network traffic. It would be more efficient if the embedding process could take place directly within the model hub.

Finally, all existing transferability metrics are limited to supervised tasks. Therefore, they are not suitable for selecting LLMs for text generation.

\section*{Ethics Statement}

Anticipating ethical concerns for this research is challenging because of the wide-ranging utility of language model selection. As fine-tuning the entire space of available language models is unsustainable and unethical in terms of climate sustainability, efficient encoder pre-selection using transferability estimation provide a positive step toward tackling this problem. \framework{} enhances the process of efficient model selection by providing a unified tool, thereby encouraging practitioners to avoid unnecessary fine-tuning.

\section*{Acknowledgements}

Max Ploner and Alan Akbik are supported by the Deutsche Forschungsgemeinschaft (DFG, German Research Foundation) under Germany’s Excellence Strategy – EXC 2002/1 “Science of Intelligence” – project number 390523135. Alan Akbik is further supported by the Deutsche Forschungsgemeinschaft (DFG, German Research Foundation) under the Emmy Noether grant ``Eidetic Representations of Natural Language'' (project number 448414230).
\bibliography{references}
\appendix

% ---------------------------------------
% ---- A. Details on Embedder Class -----
% ---------------------------------------
\section{Embedding and Pooling}
\label{appendix:embedder}

The library uses an internal \textit{Embedder} class to retrieve word or sentence-level embeddings from pre-trained language models.

For per-word classification tasks, such as NER and PoS tasks, the library supports pooling operations to extract word-level embeddings (as a single word may consist of multiple subword tokens). \textit{First} pooling uses the first subword to represent a word. \textit{Mean}, which is used by default, averages all subwords to represent a word.

For text classification tasks, \framework{} pools word embeddings to create a sentence embedding.
It supports several options: \textit{first} uses the \texttt{CLS}-token, which can be used for models that used this token during pre-training.
\textit{Mean} averages all word embeddings into a single sentence embedding.
We use \textit{mean} as the default since we aim to rank models that that were trained using different pre-training objectives and some do not use the \texttt{CLS}-token.
Using the mean representation rather than the \texttt{CLS}-token has been empirically shown to be more effective by \citet{bassignana-evidence}.
\textit{Last} uses the embedding of the last word to represent a sentence. This may be more suitable for comparing autoregressive models, although such models were not used in our experiments.

% ---------------------------------------
% ---- B. Details on GermEval18     -----
% ---------------------------------------
\section{Example using GermEval18}\label{sec:appendix-germeval}

To demonstrate the library in action, we used a dataset of offensive language identification in German tweets, available on the HuggingFace hub as '\textit{philschmid/germeval18}' through the \texttt{datasets} framework. 
To evaluate the results produced by the \framework{}, all models were fine-tuned with FLAIR using default parameters: a 5e-6 learning rate, batch size of 4, 10 epochs, along with the AdamW optimizer and a linear scheduler with a 0.1 warmup fraction. 
Table \ref{tab:demonstration-experiment} reports the average F1-micro score and standard error over five runs.

\end{document}